# Correlation between prosody and pragmatics: A case study of the discourse marker *hālā* ‘now’ in Persian

**Soleiman Ghaderi** ORCID iD: https://orcid.org/0000-0003-1608-8644
Department of Linguistics, Faculty of Foreign Languages, University of Isfahan, Isfahan, Iran

**Moloud Asakereh** ORCID iD: https://orcid.org/0009-0009-3705-5674
Department of English Language and Linguistics, Institute of English and American Studies, Faculty of Arts and Humanities, Heinrich Heine University Düsseldorf, Germany

**Kevin Tang** [*] ORCID iD: https://orcid.org/0000-0001-7382-9344
Department of English Language and Linguistics, Institute of English and American Studies, Faculty of Arts and Humanities, Heinrich Heine University Düsseldorf, Germany

**Abstract**

The Persian discourse marker *hālā* (‘now’) exhibits remarkable multifunctionality, extending far beyond its temporal adverbial role to encompass a variety of pragmatic functions. This study presents a pragmatic and acoustic analysis of *hālā* in spoken Persian, examining 267 instances from spontaneous conversations. While temporal uses were present, they were often combined with other discourse marker functions, indicating extensive multifunctionality, with 70% of tokens serving two or more pragmatic roles. Textual functions (topic shifting, signaling relationships, boundary marking, attention guidance, topic introduction, and topic emphasis) were most frequent, followed by interactive functions (turn management, listener engagement, and feedback regulation), and modal functions (epistemic stance, emotional expression, and attitudinal marking). Prosodic analysis revealed that duration and intensity are key cues for distinguishing *hālā*’s functions. Textual uses were significantly shorter, while temporal uses showed a tendency toward longer realizations. Interactive functions correlated with higher intensity, while modal functions showed a weaker tendency toward lower intensity. These findings indicate that duration and intensity are the main prosodic cues associated with functional differentiation in *hālā***,** especially in textual and interactive uses**.**

## 1 Introduction

Discourse markers (DMs) form a broad and heterogeneous class of expressions whose boundaries remain theoretically elusive. Prototypical forms such as *okay*, *now*, and *well* are widely recognized, while items like *yeah*, *so*, and *you know* occupy peripheral positions. Competing frameworks have proposed diverse accounts of these items. Schiffrin’s (1987) discourse semantic model treats DMs as coherence devices organizing talk; Blakemore’s (2002) relevance-theoretic approach argues that they encode procedural rather than conceptual meaning, guiding inferential processing; and Discourse Grammar (Heine et al., 2013), more recently extended as Interactive Grammar (Heine, 2023), categorizes them under theticals (or interactives): extra-clausal, interaction-oriented units such as interjections, vocatives, and comment clauses. Despite these divergent terminologies and theoretical differences, most accounts converge on several cross-cutting properties: multifunctionality, low conceptual content alongside procedural meaning, and being syntactically, semantically, and prosodically detached from their hosts. Functionally, DMs contribute to discourse management, stance expression, and speaker–hearer alignment (Brinton, 1996; Fraser, 1999; Heine, 2023).

[*] Corresponding author: Department of English Language and Linguistics, Institute of English and American Studies, Office: 23.21.02.102, Universitätsstr. 1, 40225 Düsseldorf, Heinrich Heine University, Germany
Email: kevin.tang@hhu.de

Cross-linguistic research confirms that DMs interact closely with prosodic structure. Studies on *now* and its equivalents, such as English *now* (Hirschberg & Litman, 1987), Hebrew *axshav* (Gonen et al., 2015), and French *alors* (Degand & Fagard, 2011), show that these DMs demonstrate how prosodic variations (e.g., boundary tone, prominence, and duration) co-signal pragmatic functions, often working in conjunction with syntactic placement and lexical choice. This evidence highlights that prosody, in addition to syntax and semantics, plays a critical role in the functional differentiation of DMs.

Within this linguistic landscape, the Persian word *hālā* ('now') emerges as a particularly compelling case study. While its etymological root is firmly in a temporal adverb, *hālā* has demonstrably expanded its functional repertoire to encompass a range of non-temporal, metatextual roles. These include, but are not limited to, marking discourse boundaries, facilitating topic shifts, and conveying subtle evaluations or expressions of mild disagreement. Few prior studies (e.g., Mohammadi, 2018; Noora, 2015) have provided the first discourse-pragmatic descriptions and explored the grammaticalization pathways of *hālā*. Nevertheless, they have not provided a comprehensive inventory of *hālā's* functions. Furthermore, they have failed to offer an integrated analysis correlating specific acoustic properties with *hālā*'s diverse discourse functions.

This study aims to bridge this research gap by adopting the theoretical lens of Discourse Grammar. By building upon functional cognitive linguistics approaches (e.g., Heine, 2023; Traugott, 2022), we examine *hālā* as a thetical element within the discourse grammar framework. The research is guided by two principal objectives: (1) to identify and categorize the primary thetical functions of *hālā* as evidenced in contemporary colloquial Persian speech, and (2) to investigate the correlations between key acoustic parameters of prosody (specifically duration, pitch, and intensity) and these functions. Drawing inspiration from theories positing that frequently used lexical items undergoing grammaticalization often experience phonetic and prosodic restructuring (Bybee, 2003; Lohmann, 2017), we hypothesize that *hālā* will exhibit measurable acoustic distinctions when deployed in its temporal-adverbial capacity versus its discourse marker functions. The analysis offers a brief examination of syntagmatic and paradigmatic dimensions, focusing specifically on linear order and a select number of collocations. This approach also captures the dynamic interplay of grammaticalization (Heine et al., 2020) and constructionalization (Traugott, 2022). Additionally, we consider how indexicality and prosodic re-profiling contribute to the emergence of new pragmatic and interactional functions (Grosz & Michelas, 2022; Schiffrin, 1987).

The subsequent sections are structured as follows: Section 2 reviews theoretical and cross-linguistic perspectives on lexical items, primarily 'now', and their evolution into discourse marking and associated prosody. Section 3 details the methodological design and corpus data. Section 4 presents the functional and prosodic analyses of *hālā*. Finally, Section 5 discusses the broader implications of the findings for theories of thetical evolution and multifunctionality in DMs.

## 2 Multifunctionality and prosodic variation in discourse markers

Studies demonstrate that DMs often possess a range of pragmatic functions, with prosodic variations acting as primary cues to these roles. To understand this interaction, we will first review the pragmatic dimensions of DMs (Section 2.1) and then explore their prosodic characteristics (Section 2.2). Finally, we will review previous research on 'now' and Persian discourse markers (Section 2.3).

### *2.1 Pragmatics of DMs*

The study of DMs boasts a rich history in linguistic research, evolving through various classification systems aimed at understanding their multifaceted roles. Linguistic inquiry has focused on the specific contributions of DMs to discourse coherence and organization. Schiffrin's (1987) work, for instance, analyzed DMs through the lens of discourse structure, linking them to planes such as ideational, action, participation, exchange, and information states. Fraser (1999) further refined this by concentrating on pragmatic functions, proposing categories like topic, contrastive, and elaborative markers. More recently, Hyland's (2007) metadiscourse model distinguished between interactive functions (e.g., transitions, frame markers) and interactional functions (e.g., hedges, boosters), emphasizing how these markers manage the flow of information and the writer-hearer relationship.

While these frameworks offer valuable perspectives, Bernd Heine's Discourse Grammar provides a particularly comprehensive approach that integrates the diachronic process of grammaticalization with the synchronic analysis of DM functions. This framework posits that DMs, as thetical expressions, operate at the macrostructure level of communication. The macrostructure level is fundamentally metadiscursive, where theticals relate the text to the broader situation of discourse, which encompasses text organization, speaker attitudes (subjectivity), and speaker-hearer interaction (intersubjectivity). This contrasts with the microstructure (sentence grammar), which deals with the propositional content and structural elements of clauses and sentences. Central to Discourse Grammar is the concept of cooptation, the process by which a chunk of sentence grammar (microstructure), such as a clause, phrase, or word, is deployed for thetical grammar (macrostructure). Following cooptation, these units, initially conceptual theticals, can undergo grammaticalization to become established, multifunctional, polysemous DMs, acquiring a range of distinct pragmatic functions (Heine et al., 2013, 2020).

A clear illustration of this phenomenon can be observed in the Persian word *hālā*:

(1)
A: *hālā šomā mi-r-in?*
now you IPFV-go.PRS-2SG
'Are you going now (at the present time)?'
B: *hālā šomā mi-r-in?*
now you IPFV-go.PRS-2SG
'Now (finally), will you go there?'

In example (1a), *hālā* functions as a temporal adverb within sentence grammar, semantically, syntactically, and prosodically integrated into the sentence to structure a specific point in time. In contrast, in example (1b), *hālā* operates as a thetical element within discourse grammar. Such discourse-level *hālā* instances are typically invariable and syntactically, semantically, and prosodically independent of the host utterance. Their function is metatextual, carrying a pragmatic scope that extends to the entire utterance or even the broader discourse context. The independence enables DMs to perform multiple functions:

organizing discourse (textual functions), expressing speaker attitude (modal functions), or managing speaker-hearer interaction (interactive functions). This framework underscores that seemingly simple items identified as DMs, such as *hālā* and *now*, possess a complex grammatical life, evolving from propositional elements to crucial tools for navigating communication. This pathway frequently results in the emergence of homophonous or near-homophonous linguistic forms as seen for *hālā* in examples (1a) and (1b).

## *2.2 Prosody of DMs*

Disambiguating the functions of DMs often hinges on prosodic cues. As illustrated in examples (1a) and (1b), these cues, such as a pause, distinguish between a temporal reference and a DM. Linguistic research on DMs in this area frequently delves into their phonological characteristics, including pauses, phonetic reduction, and intonation (see Brinton, 1996; Hirschberg & Litman, 1987, 1993; Romero-Trillo, 2015; Schiffrin, 1987; Tabor & Traugott, 1998). Pauses may precede a DM, particularly when it initiates an intonational unit, and a following pause is also common, potentially due to syntactic detachment (Schiffrin, 1987). Traugott (1995, p. 60) posits that these pauses can form "an independent breath unit carrying a special intonation and stress pattern." However, the prosodic independence of DMs is not absolute; for instance, Hirschberg and Litman (1993) found that the English DM *well* exhibited prosodic independence in only 50% of cases. Similarly, Dér and Markó (2010), in their pilot study of Hungarian discourse markers, found that discourse markers were not consistently flanked by pauses, suggesting that prosodic independence is not a necessary feature of DM status across languages.

Phonetic erosion, or phonological reduction, is widely recognized as a fundamental process in grammaticalization, i.e., the gradual transition from lexical to grammatical forms, or from a less grammatical to a more grammatical status. This process frequently co-occurs with semantic bleaching, meaning extension, and decategorization (Heine & Kuteva, 2007, pp. 33–46). While the direct link between semantic bleaching and phonetic reduction is not always clearly defined (Campbell, 2001), reduced prosodic salience can emerge from the inherent loss of semantic weight during grammaticalization (Wichmann, 2011). Such reduction may be driven by increased frequency of use and the tendency for grammatical elements to occupy less stressed positions within utterances (Heine et al., 1991, p. 214). It is important to note, however, that phonological reduction is neither a necessary nor a sufficient condition for grammaticalization (Lessau, 1994, p. 263).

In the same domain, the principle of the probabilistic reduction hypothesis (Jurafsky, 2003) suggests that high-frequency words are more prone to reduction than their low-frequency counterparts. Further elaborating on this, Tang and Bennett (2018) argue that high contextual probability, i.e. the likelihood of a linguistic unit appearing in a specific local environment, diminishes the phonetic salience of morphemes, words, and other units. Of course, this reduction can also occur through mechanisms such as articulatory routinization (Bybee, 2001, p. 78), decreasing the informational contribution of the grammaticalized elements attributed to meaning generalization (Bybee, 2003), and highlighting iconicity (Dehé & Stathi, 2016). Whatever the cause, the specific acoustic details of this reduction become integrated into the mental representation of the word (Tang & Shaw, 2021).

The phonetic and acoustic realization indicates that pairs which appear homophonous, such as the temporal and DM uses of *hālā*, may not reveal identical phonetic features such as duration, intonation and pitch. Even within the domain of thetical grammar, instances such as *hālā* can exhibit distinct phonetic characteristics depending on their specific thetical function, suggesting a nuanced relationship between form and meaning. This is also motivated by recent findings demonstrating that prosody systematically encodes discourse-level pragmatic distinctions, such as subjective vs. objective causality (Hu et al., 2025) and different pragmatic functions of wh-questions (Baltazani et al., 2020).

Although an initial acoustic impression suggests that seemingly homophonous pairs, such as temporal and DM *hālā*, may differ in duration, intonation, and pitch, and that these prosodic distinctions may extend to specific thetical functions, these potential patterns have not been systematically examined. To move beyond intuition, the present study provides the first rigorous, quantitative analysis of this form-meaning interplay, motivated by recent work demonstrating that prosody systematically encodes pragmatic functions (e.g., Hu et al., 2025, on subjective vs. objective causality; Baltazani et al., 2020, on the prosody of wh-questions).

### *2.3 Studies on 'now' and Persian discourse markers*

Drawing on this theoretical background, this section reviews key studies that illuminate the multifunctionality and prosodic variation of DMs, starting with cross-linguistic examples and then delving into Persian research. Degand and Fagard (2011) showed that the French DM *alors* has similarly evolved from a temporal adverbial to a temporal connective, to a causal-conditional, and to a DM. Hirschberg and Litman (1987, 1993) established that prosody is crucial for distinguishing English *now* as a DM from its temporal use. Their findings revealed that DM *now* tends to be prosodically independent, initially placed, and exhibits lower pitch accent patterns compared to temporal *now*, suggesting phonetic reduction associated with grammaticalization.

Gonen et al. (2015) acoustically analyzed Hebrew *axshav* ('now') in spoken conversation, finding that over a third of tokens functioned as DMs for segmentation, highlighting information, and holding the floor. Acoustically, DM *axshav* differed in intonation and was significantly shorter than temporal *axshav*, reinforcing the role of prosody in function differentiation.

Functional studies on Persian DMs have highlighted their multifunctionality. Noora (2015), using a discourse-pragmatic approach, analyzed *hālā* through its subjective, textual, intersubjective, and presuppositional roles, linking this to its grammaticalization. Ghaderi (2022) detailed the diverse roles of *nā* 'no' as a negative/affirmative response, inhibition, or tag question, and its procedural functions in discourse management. Ghaderi (2024) categorized Persian response signals, highlighting their roles in (dis)confirmation, acceptance/rejection, and backchanneling.

Regarding the prosody-function relationship, Ghaderi (2024) also observed cross-linguistic phonological patterns in response signals (e.g., /ʔ/ or /h/ for positive tokens, clicks for negative ones). Kazemian et al. (2025) found that the pragmatic functions of *bebin* 'look' are distinguished primarily by duration: more grammaticalized uses (attention signal, discourse marker) are significantly shorter, while the interjection use is longest, with no significant differences in F0 or intensity. More broadly, recent work on the prosodic realization of focus in Persian (Taheri-Ardali et al., 2025) have established that duration, intensity, and pitch are key acoustic cues for information-structural distinctions. Ghaderi (2021) noted that the duration and intonation of *xob* 'well' vary with its function (e.g., longer duration with rising intonation for consensus-seeking, shorter with falling for acceptance). Mohammadi (2018) examined a few functions of *xob*, *āxe*, and *hālā*, but conducted instrumental prosodic analysis only for *xob*: short duration + falling pitch signals cognitive involvement (comprehension), while longer duration + rising pitch signals affective involvement (emotional engagement). For *hālā*, she described a few discoursal roles as a focus anchor for discourse structure and closure and its attitudinal function of temporal dismissal, without prosodic analysis.

The present study investigates the Persian discourse marker *hālā*, aiming to clarify its pragmatic functions as well as their associated prosodic information (duration, F0, intensity). In the following section, we outline the data and methodological procedures used to address these questions.

# 3 Methodology

The present study adopts a multi-layered methodological design to examine both the pragmatic functions and acoustic correlates of *hālā*. Drawing on spontaneous conversational data, we apply manual functional coding alongside systematic acoustic analysis to capture the form-meaning interplay at the discourse and phonetic levels.

## *3.1 Theoretical framework*

Our analysis draws upon the dual-model framework of Discourse Grammar (Heine et al., 2013), more recently expanded as Interactive Grammar (Heine, 2023). This framework challenges strictly unidirectional views of DMs, such as a single form-single function approach or homonymy where unrelated meanings are assigned to the same form. Instead, it draws a distinction between a word's original conceptual meaning functions at the microstructural level (in this case, the temporal sense of 'now' in the adverb *hālā*) and its emergent pragmatic functions at the macrostructural level that have arisen through processes of cooptation and grammaticalization. This framework predicts that some of these functions may be associated with different phonetic realizations.

## *3.2 Data and corpus*

Consistent with the Discourse Grammar emphasis on corpus-based analysis, this study utilizes empirical data extracted from the Farsi (Persian) subset of the Multi-Language Conversational Telephone Speech Corpus (Linguistic Data Consortium [LDC], 2011). This corpus predominantly comprises naturalistic data from semi-formal and informal interactions in Contemporary Iranian Standard Persian, specifically the Tehran dialect, making it highly suitable for analyzing the recurring functions of the DM *hālā*. The recordings consist of spontaneous telephone conversations between speakers and are provided in a dual-channel format, corresponding to the caller and the callee, which allows for speaker-specific analysis.

The full Farsi subset of the corpus comprises 100 conversational telephone recordings. To ensure a representative and manageable dataset for in-depth analysis, 65 files were selected from this larger set**,** following principles of corpus representativeness (Biber, 1993). From the selected data, a total of 267 tokens of *hālā* were identified and extracted for analysis, drawn from a diverse set of speakers to mitigate subgroup bias. A minimum of 10 tokens for each sub-function of the DM *hālā* was considered suitable for a comprehensive exploration of its functions and prosodic features, in line with established methodological practices in discourse marker research (see, e.g., Crible & Degand, 2019). As shown in Table 1, the dataset reflects the multifunctional nature of *hālā*, including both overall functional distributions and patterns of functional overlap.

Table 1. Approximate functional distribution and overlap of functions in 267 tokens of *hālā*

| Panel | Category | N |
|---|---|---|
| **A. Overall functional categories** | Textual | 211 |
| | Temporal | 127 |
| | Interactive | 100 |
| | Modal | 92 |
| **B. Functional overlap in tokens** | Tokens with single function (Non-overlapping) | 82 |
| | Tokens with two functions | 113 |
| | Tokens with three functions | 66 |
| | Tokens with four functions | 6 |

The extraction of tokens and their temporal boundaries were based on forced alignment outputs obtained using the Montreal Forced Aligner (MFA), in combination with manually inspected TextGrid annotations. This procedure ensured precise temporal localization of each token and is described in the next section.

## *3.3 Analytical approach*

Our analysis employs a mixed-methods approach, combining qualitative and quantitative methods. We first qualitatively examine how *hālā* interacts with specific discourse contexts. This involves identifying its situational functions and exploring common collocations observed across these functions. This approach aligns with corpus-based analysis that emphasizes real-world language use. The qualitative findings are complemented by quantitative measures to assess the frequency of *hālā* across its defined functions and their overlaps. This integrated approach synthesizes inductive data observation with the deductive testing of the Discourse Grammar framework. In addition, prosodic analysis is performed quantitatively and qualitatively to examine acoustic correlates of the identified functions; this is described in detail in Sections 3.6, 3.7, 4.2, and 4.3.

Diachronically in line with Hansen's (2006) concept of 'diachronic responsibility,' this synchronic study provides evidence for the grammaticalization of *hālā* from its adverbial root. However, the precise diachronic emergence of *hālā*'s functions is reserved for future diachronic studies. This methodological stance underscores the necessity of aligning synchronic descriptions with historical evidence; thereby avoiding assumptions about meaning development that are not supported by historical data.

## *3.4 Transcription and time alignment*

The acoustic data were automatically transcribed using *gemini-2.0-pro-exp-02-05*, an experimental public preview version of Gemini 2.0 Pro (Google DeepMind, 2025), accessed via Google AI Studio (https://aistudio.google.com/). The acoustic recordings were transcribed one at a time. This specific model was chosen after a manual evaluation of transcription quality on a random subset of the recordings (approximately 30), comparing the Gemini model with three other ASR models: WhisperX (Bain et al., 2023), Google ASR accessed via the BAS CLARIN web services (Kisler et al., 2017), as well as VOSK (model: fa-0.24) via the Subtitle Edit software (version: 4.0.11). The Gemini model yielded the most accurate transcriptions among these systems.

The Gemini transcripts were then used to identify utterances that contain the lexical item *hālā*. The transcriptions of these utterances, together with their surrounding context, were manually checked and corrected by the first author, who is a native speaker of Persian, to ensure accuracy.

Time-aligned word-level and phone-level boundaries were obtained using the Montreal Forced Aligner (MFA; version 2.2.17). Because no pretrained Persian acoustic model was available, we developed a custom Persian acoustic model and applied it to the dataset. The resulting alignments were manually corrected by the second author, who is a native speaker of Persian with phonetic training, in Praat (version 6.4.28; Boersma & Weenink, 2024), with particular attention to regions affected by coarticulation, reduction, and background noise, in order to ensure precise segmentation.

For a detailed description of the development and training of the Persian forced alignment model, see Appendix A.

All instances of the lexical item *hālā* were identified based on the hand-corrected word-level segmentations. Utterances containing *hālā* were segmented using the surrounding pauses. Following the practice of Seyfarth (2014), an utterance is defined as a stretch of speech produced by a single speaker and delimited by pauses around the target token. For a detailed description of the utterance segmentation, see Appendices B and C.

## *3.5 Functional coding and reliability*

At this stage of the analysis, the pragmatic function of each *hālā* token was manually annotated. We classified the range of functions observed in the data, including its use as a temporal adverbial as well as its discourse-pragmatic roles. These were then grouped into four main domains: temporal, interactive, modal, and textual.

We analyzed each token of *hālā* in relation to its discourse context and assigned one or more functional labels depending on how it contributed to the utterance and the surrounding conversation. Because DMs are often multifunctional, the categories were not treated as mutually exclusive, and each token could receive more than one functional label.

The functional analysis was based on manual, qualitative coding, employing a dynamic and iterative process between theoretical frameworks and empirical data. Classifying the polysemous discursive functions of *hālā* presented a significant challenge. To ensure the reliability of this classification, two trained linguists independently coded the data. They achieved 90% agreement on the primary functional categories (textual, interactive, and modal) and subsequently 75% on finer sub-functions. This level of inter-annotator agreement is considered a strong validation, guided by the principle of methodological minimalism (Hansen, 2006). This approach favored explaining occurrences within existing (sub)functional categories and contexts over introducing new, potentially overgenerated categories.

This approach makes it possible to examine not only how individual functions are distributed, but also how they overlap, which reflects the multifunctional nature of *hālā* in spontaneous speech. Representative examples in this paper are glossed and translated into English following the Leipzig Glossing Rules (Comrie, Haspelmath, & Bickel, 2008; see Appendix D for abbreviations).

## *3.6 Acoustic measurements*

In addition to functional annotation, we extracted acoustic measurements for each token of *hālā* using the aligned TextGrid files. All measurements were carried out in Praat via the Parselmouth Python library (version 0.4.5; Jadoul et al., 2018).

For each token, we calculated word-level duration based on the onset and offset times in the hand-corrected alignments. Pitch (F0) and intensity were also measured. Pitch was estimated using an autocorrelation method, with a floor of 75 Hz and a ceiling of 600 Hz. From the pitch contour, summary measures including mean pitch, maximum pitch, and pitch range were extracted. Intensity was measured in decibels (dB), and summary measures including mean intensity, maximum intensity, and intensity range were calculated.

To control for the potential effect of speaking rate on duration, pitch and intensity, we also measured contextual speech rate. For each token, two intervals were defined: a pre-*hālā* window extending from the beginning of the utterance to the onset of the token, and a post-*hālā* window extending from the token's offset to the end of the utterance. Within each window, syllables were approximated by counting vowel-labeled phones in the aligned phone tier. Speech rate was then calculated as the number of vowel segments divided by interval duration (syllables per second).

Finally, we combined the acoustic measurements and functional annotations into a single token-level dataset, which formed the basis for all subsequent statistical analyses, including both the main acoustic models and the additional mediation models**.**

### *3.7 Statistical analysis*

To examine the relationship between pragmatic function and the acoustic realization of *hālā*, we fitted linear mixed-effects models using the lme4 package (version 1.1-37; Bates et al., 2015) in R (version 4.5.0; R Core Team, 2025), using RStudio (version 2025.09.1+401; Posit team, 2025). All analyses were carried out at the token level, with each occurrence of *hālā* treated as an individual observation. To account for variation between speakers, we included speaker as a random intercept in all models.

All functional predictors were coded as binary variables indicating whether each functional domain was present or absent for each token. These variables were then contrast-coded for the statistical models, with presence coded as +0.5 and absence as −0.5. This makes the coefficients easier to interpret relative to the overall mean. Because the functional categories were not exclusive and a single token could serve multiple functions, we entered all functional predictors (temporal, textual, modal, and interactive) simultaneously in the initial models. This approach allowed us to estimate the effect of each function while controlling for overlap between them.

We analyzed three word-level acoustic measures separately: word duration (in milliseconds), maximum pitch (in Hz), and maximum intensity (in dB). Duration was $\log_{10}$-transformed to reduce skewness and make the distribution more symmetric; in the raw data, duration showed moderate positive skewness (skewness = 1.44), which was substantially reduced after $\log_{10}$ transformation (skewness = 0.14). Maximum pitch was also $\log_{10}$-transformed after comparison with alternative scales (e.g., raw Hz, Mel, Bark); among these, the $\log_{10}$ transformation yielded the least skewed distribution (skewness = 0.12) and was therefore selected for the statistical analysis. Intensity was analyzed on its original scale. We included pre- and post-*hālā* speech rate as z-transformed covariates in all models to control for local differences in speaking rate. Z-transformation helped with model convergence and made it easier to compare coefficients across predictors with different numerical scales.

For each dependent variable, the initial model had the following structure:

DV ~ Pre_SpeechRate + Post_SpeechRate + Temporal + Textual + Modal + Interactive + (1 | Speaker)

where DV was duration, pitch, or intensity. Model selection was carried out using a backward nested model comparison procedure. Starting from the full model for each dependent variable, we removed one functional predictor at a time and compared the reduced model with the fuller model using likelihood ratio tests under maximum likelihood estimation (REML = FALSE). Functional predictors whose removal did not significantly worsen model fit (p-value > 0.10) were excluded from the models. The speech-rate covariates were retained in all models. The resulting reduced models were treated as the final fixed-effects structures and were then refitted using restricted maximum likelihood (REML = TRUE) to obtain the reported parameter estimates. Potential collinearity among predictors was evaluated using correlation inspection and variance inflation factors; all VIF values were close to 1, which indicate no serious issues of collinearity.

In addition to these models, a series of mediation analyses was conducted to examine whether the functional effects observed in one acoustic model could be explained by another acoustic dimension. In these models, pitch and intensity were added to the duration models, duration and intensity were added to the pitch models, and duration and pitch were added to the intensity models. These mediation-style models used the same random-effects structure and the same speech-rate covariates as the main analyses.

All models underwent the process of model criticism. For each model, residuals were examined using Q-Q plots and residuals-versus-fitted plots. Data points that were 2.5 standard deviations above or below the mean residual value were excluded. Because trimming was carried out separately for each main and secondary model, the number of observations varies slightly across models. As in the main analyses, the trimmed models are the ones reported and interpreted in the Results section.

# 4 Results

In this part, we first present the functional analysis of *hālā* in Section 4.1, followed by the prosodic analysis in Sections 4.2 and 4.3.

## *4.1 Functional analysis of hālā in Persian discourse*

Across the 65 files analyzed, a total of 267 tokens of *hālā* were identified. Individual speakers varied considerably in their use of the marker, with frequencies ranging from zero to thirty occurrences in a file (each file approximately 1,324 words in length, shared between two speakers), reflecting substantial inter-speaker variation. This variability also underscores the dynamic and context-dependent nature of this DM. *Hālā* (meaning 'now') functions both as a deictic temporal adverbial with conceptual meaning (Section 4.1.1) and as a DM performing various procedural functions (Sections 4.1.2–4.1.4). In this section, we examine different functions of *hālā* across four main domains: temporal, interactive, modal, and textual. Notably, these functions often overlap, reflecting the multifunctional nature of *hālā*. In Sections 4.2 and 4.3, we examine how these functions are reflected in the prosodic realization of *hālā*. Please see Appendix D for a breakdown of the abbreviations and phonetic notations used below.

### *4.1.1 Temporal adverbial functions of hālā*

Before examining its DM functions, it is necessary to briefly consider *hālā*'s original temporal role. As a deictic temporal adverb, *hālā* conceptually anchors events to the present moment, as in example (2):

(2)

| *če* | *kard-in?* | *va* | *hālā* | *dār-id* | *či-kār* | *mi-kon-id?* |
|---|---|---|---|---|---|---|
| what | do.PST-2PL? | and | now | have.AUX.PRS-2PL | what-work | IPFV-do.PRS-2PL? |

‘What did you do? And now what are you doing?’

The temporal word *hālā* can be integrated into various constructions, imbuing them with specific meanings. For instance, the adjective ‘*hālāʔi*’, ‘nowadays’ pertains to contemporary things or individuals. The reduplicated adverb *hālā-hālā-hā,* literally ‘now-now-plural’ signifies ‘for a long, indefinite time.’ Another construction, *hālā* $[X]_{V.}$ *va. key* $[X]_{V.}$ , literally meaning ‘now V. and when V.’, where one verb in the pair is negated, denotes the continuity of an action. An example is *hālā nazan o kei bezan*, which refers to the ongoing act of hitting someone. Similarly, *hālā* can form an intriguing paradoxical collocation with *baʔd* (later) in the phrase *hālā baʔd* ‘now later’. This expression indicates a postponement, suggesting that something is not being addressed or is being dismissed at the present moment and may be deferred. In this case, *hālā* appears to shift from its role as a deictic term for the present time to functioning as a DM, while *baʔd* retains its meaning of ‘later.’

### *4.1.2 Interactive functions of hālā*

This section categorizes and explains the interactive roles of *hālā* into turn management, feedback regulation, and listener engagement. Interactive functions help speakers maintain smooth communication and ensure listener involvement.

*4.1.2.1* ***Hālā as a turn-management marker.*** *Hālā* functions as a turn-management marker, organizing speaker transitions in conversation. It signals the start, holding, or yielding of a turn, regulating conversational flow and minimizing overlaps.

(3)

A: *Man dige sohbat ne-mi-kon-am*
I anymore talk NEG-IPFV-do.PRS-1SG
guš mi-d-am be sohbat-hā-ye šomā.
ear IPFV-give.PRS-1SG to talke-PL-EZ your.

B: *hālā šomā har mowqeʔ ke xāst-id*
Now you every occasion that want.PST-2PL
*mi-tun-id sohbat-e man=o qatʔ kon-id.*
IPFV-be.able.PRS-2PL talk-EZ my=OM cut.off SBJV.do.PRS-2PL.

A: ‘I won’t speak anymore and listen to your speech.’

B: ‘Now, you can interrupt me anytime if I start rambling too much.’

Here, *hālā* in (3B) signals turn-taking by another participant. This aligns with Sacks et al.’s (1974) model of turn-taking, where markers mitigate overlaps and maintain orderly interaction.

*Hālā* can serve as a turn-holding device, (as in examples 7 and 8), marking hesitation while the speaker organizes their thoughts. This use prevents interruption and allows the speaker to maintain control of the conversational floor. While *hālā* frequently signals turn initiation or retention, it also occurs in turn-yielding contexts, most notably when prompting the listener’s response (see examples 4-6).

*4.1.2.2* ***Hālā as a feedback-regulation marker.*** *Hālā* may function as a feedback marker, used to elicit or confirm understanding between speakers. In the data, *hālā* often appears with epistemic hedges such as *nemidunam* ("I don't know"), soliciting acknowledgment or agreement feedback.

(4)

*... be hadafāsh resid-e Ø. hālā man ne-mi-dun-am*
… to goals-his reach-PP be.AUX.PRS.3SG Now I NEG- IPFV-know.PRS-1SG
*šomā tā če had in harf rā bāvar dār-id.*
you to what extent this statement OM belief have.PRS-2PL.
[He just emitted that energy from himself and] 'achieved his goal. Now, I don't know to what extent you believe this statement.'

In example (4), *hālā* introduces a confirmation-seeking move, inviting the listener to validate the speaker's claim about energy originating from thoughts related to achieving goals. The collocation with *nemidunam* mitigates potential face threats that may arise from imposing an idea or facing possible disagreement, reflecting politeness strategies (see Brown & Levinson, 1987; Schegloff, 2007, pp. 58–81).

(5)

A: *…Hālā ne-mi-dun-am če-qadr dige bāyad*
…Now NEG-IPFV-know.PRS-1SG how-much other must
*harf be-zan-am vali-*
talk SBJV-make.PRS-1SG but-
B: *zamān dārim hanuz sohbat bo-kon-im bā ham.dige.*
time have.PRS-1PL still talk SBJV-do.PRS-1PL with each.other.
A: '[I'm trying to travel the world and enjoy life for now.] 'Now, I don't know how long I should talk, but...'
B: 'We still have time to talk together.'

In (5A), *hālā* acts as a topic probe, testing the listener's readiness to continue the conversation. This aligns with Schiffrin's (1987) notion of feedback markers that scaffold mutual understanding. Overall, in this role *hālā* helps speakers gauge topic appropriateness, invite listener responses, and regulate conversational flow. By this engaging the interlocutor and facilitating feedback-based alignment, *hālā* supports ongoing dialogic coherence.

*4.1.2.3* ***Hālā as a listener-engagement marker.*** *Hālā* functions as an engagement marker, drawing the listener into the conversation more deeply than a simple request for confirmation. It not only directs the listener's attention but also mitigates potential face threats, fostering collaborative dialogue that makes the interlocutor feel included, valued, and often prompted to participate or act.

(6)

*Fārsi sohbat bo-kon-id. Hālā agar dust dār-id*
Persian talk SBJV-do.PRS-2PL Now if like have.PRS-2PL
*yek xātere=i barāye mā tarif kon-in*
a memory=INDF for us tell SBJV-do.PRS-2PL
*dust dār-in ye šer=i be-xon-in.*
like have.PRS-2PL a poem=INDF SBJV-read.PRS-2PL.
'Speak in Persian. Now, if you'd like, share a memory with us or recite a poem.'

As a metadiscursive cue, *hālā* signals the speaker's intention to cede topic selection to the listener, scaffolding participatory dialogue. This is often achieved through indirect requests like *age dust dāri* ("if you'd like") or *ʔāyā mituni* ("if you can"). This Persian-specific pragmatic indirectness (Koutlaki, 2002), which prioritizes listener autonomy and face preservation, mirrors politeness strategies that reduce imposition (Brown & Levinson, 1987; Fraser, 1990) by framing requests as optional and using epistemic mitigation (e.g., with *nemidunam*) (see also Sections 4.1.3.3 and 4.1.2.2). While *hālā*'s interactive functions in managing turns and eliciting feedback parallel those of English *now*, its specific collocations and cultural emphasis on politeness through indirectness highlight unique Persian pragmatic norms.

### *4.1.3 Modal functions of hālā*

In Persian, the word *hālā* serves multiple modal functions, including expressing attitudes, emotions, and epistemic stance. Below, we categorize the functions and offer a brief analysis for each category.

*4.1.3.1* ***Hālā as an attitude marker.*** *Hālā* is often used to express the speaker's attitude, such as approval, disapproval, or politeness. More specifically, as an attitude marker, it reflects the speaker's evaluative response to the content.

(7)

*Dorost=e, vaqeʔan ʔelm xeyli pišraft kard-e Ø...*
right=COP.PRS.3SG, really science very advanced do-PP be.AUX.PRS.3SG
*hālā ham čiz-ā=ye xub=i dāšt-e Ø.*
now also thing-PL=EZ good=INDF have-PP be.AUX.PRS.3SG
*ham čiz-hā=ye bad=i.*
also things-PL=EZ bad=INDF.
'It's true, science has really advanced, and now it has had both good and bad things.'

In this example, *hālā* is used to present a balanced view, reflecting the speaker's attitude toward the dual nature of scientific progress. It functions as a DM to introduce a subjective evaluation.

*4.1.3.2* ***Hālā as an emotional marker.*** *Hālā* can also function as an emotional marker, conveying the speaker's affective reactions such as excitement, frustration, or concern.

(8)

*Agar ke mi-tunest-am bā hame ...*
in.case that IPFV-be.able.PST-1SG with all
*be-r-im be yek ǰāy=e garm=tar vaqeʔan ideāl bud*
SBJV-go.PRS-1PL to a place=EZ warm=er really ideal be.PST.3SG,
*vali xob hālā ke dige ne-mi-š-e.*
but well now that anymore NEG-IPFV-become.PRS-3SG.
'If I could move with all [my friends and sisters] and go to a warmer place, it would really be ideal, but well, now it's not possible anymore.'

Here, *hālā* conveys resignation while accepting a situation that cannot be changed, reflecting the speaker's emotional state of frustration.

*4.1.3.3* ***Hālā as an epistemic marker.*** *Hālā* can also function as an epistemic marker, signaling the speaker's degree of certainty, doubt, or possibility. Below, *hālā* conveys a mix of contentment and uncertainty about the period of time, reflecting the speaker's epistemic stance.

(9)

*...hālā dige tā key inǰā mundegār.beš-im xodā dān-ad*
… now anymore until when here SBJV.stay.PRS-1PL God know.PRS-3SG
[In short, it has become very beautiful here...] 'Now, how long we will stay here, God knows'

The common collocation of *hālā* with *nemidunam 'I don't know'* as in examples (4) and (5), signals an epistemic dimension of hesitation or uncertainty. As a multifunctional item, this may simultaneously prompts the listener to evaluate the statement's reliability and implicitly invites feedback (see Section 4.1.2.2).

The data indicate that *hālā* frequently functions as an approximation marker, often linked to epistemic modality by signaling an estimation or uncertainty regarding forthcoming or recalled information. In example (10), *hālā* conveys approximation to a specific bodily posture (e.g., "bowing") used to show respect. In example (11), it denotes an indeterminate past time span (several years), reflecting the speaker's limited confidence or uncertainty about the temporal reference.

(10)

*Mā vaqti salām.aleyk mi-kon-im xam mi-š-im*
We when greeting IPFV-do.PRS-2PL bend IPFV-become.PRS-1PL
*tā ʔun hālat hālā korneš kard-an.*
until that state now bowing do-INF.
'When we say hello, we bend now to the state of bowing.'

(11)

*hālā čand sāl=i tu yek ǰā=ye kučik=i ,*
Now several year=INDF in a place=EZ small=INDF,
*sumeʔe=ye kučik=i zendegi mi-kard-e Ø*
monastery=EZ small=INDF life IPFV-do-PP be.AUX.PRS.3SG .
'Now, for several years, he lived in a small place, a small monastery.'

## 4.1.4 *Textual functions of hālā*

According to data analysis, the Persian DM *hālā* has two primary textual roles: thematic, involving topic management; and structural, marking discourse boundaries and relationships.

### *4.1.4.1 Thematic functions of* **hālā** *in Persian discourse*

In Persian discourse, *hālā* exhibits multiple thematic functions, which can be categorized into topic introduction, topic shift, topic emphasis, and attention guidance.

***4.1.4.1.1 Hālā as a topic-introducing marker.*** *Hālā* functions as a frame marker, signaling the initiation of a new topic. It creates boundaries between discourse segments, preparing listeners for upcoming new content.

(12)

*hālā šomā dust dār-in ke ye xātere bar=ām taʔrif.kon-id?*
Now you like have.AUX.PRS-2PL that a memory for=PC.1SG SBJV.tell.PRS-2SG?

[I have about seven or eight minutes to record your lovely voice.] ‘Now, would you like to share a memory with me?’

Here, *hālā* marks a transition from the speaker’s preparatory statement (recording time) to introduce a new topic for speaking, i.e., a direct request for sharing a memory. Besides, it signals a discourse boundary, reframing the interaction toward a new (interrogative) speech act.

***4.1.4.1.2 Hālā as a topic-shift marker.*** *Hālā* acts as a topic shift marker, enabling speakers to pivot between topics or revisit prior ones. It ensures coherence by explicitly marking transitions from one (sub)topic to another (sub)topic, helping to manage the flow of discourse.

(13)

*hālā bargard-im be ʔinke be man be-gu-Ø*
now return.PRS-1PL to the.fact.that to me IMP-give.PRS-2SG
*ʔaqid=at rāǰe be dust či=e?*
belief=PC.2SG about to friend what=COP.PRS.3SG?
‘Now, let’s return to you telling me your opinion on friendship.’

In (13), *hālā* facilitates topic reinstatement, redirecting focus to a prior subject, i.e., friendship. This aligns with Fraser’s (1999) model of topic management, where markers signal shifts for dialogic coherence.

***4.1.4.1.3 Hālā as an emphatic marker.*** *Hālā* serves as an emphatic marker, underscoring the salience of a topic. It amplifies the relevance of propositions, often to highlight comparison or urgency.

(14)

*...bāyad bāš-e, to bāyad dāšt-e baš-i,*
...must SBJV.be.PRS-2PL, you must have-PP SBJV.be.AUX.PRS-2SG,
*yā man bāyad dāšte bāš-am*
or I must have SBJV.be.AUX.PRS-1SG
*va az ʔun hālā xeyli mohem-tar ʔinke*
and than that now very important-er the.fact.that
*be-mahz-e ʔin-ke kučik-tarin moškel=i piš.mi.yā-d,*
as.soon.as the.fact.that small-est problem=INDF IPFV.happen.PRS-3SG,
*in moškel manšaʔ=e talāq mi-š-e.*
this problem cause=EZ divorce IPFV-become.PRS-3SG
‘...it must exist (you should have it, or I should) but now, more importantly, the moment the smallest problem arises, it leads to divorce.’

*Hālā* here introduces a contrastive focus, emphasizing the significance of the subsequent issue (i.e., the inability of couples to resolve problems, which leads to divorce) in comparison with the previously discussed monetary issues. This mirrors Blakemore’s (2002) emphasis markers, which guide listeners to interpret statements as critical or contrastive.

***4.1.4.1.4 Hālā as an attention-guiding marker.*** *Hālā* operates as an attention-guiding marker, directing listeners to critical points. It acts as a pragmatic signpost, ensuring engagement and focus.

(15)

*hālā čerā be-sāz-im baʔd be-xāh-im tamiz kon-im?*
now why SBJV-make.PRS-1PL then SBJV-want.PRS-1PL clean SBJV.do.PRS-1PL?
'Now, why must we build something that we should clean later?'

In example (15), *hālā* functions as a metadiscursive cue, directing the listener's attention to the topic at hand and ensuring they follow the speaker's line of thought. Here, the speaker challenges a premise through a rhetorical question, guiding attention to the issue of why one would build something harmful to the environment and later clean up its remnants. Such markers serve to scaffold listener comprehension, aiding in the organization and interpretation of discourse.

*Hālā* may function as an attenuated 'no,' mitigating the preference for continuing a turn (see Schegloff, 2007, pp. 64–97). In the same domain, though rare, it may mark a speaker's withdrawal from a new coming topic rather than engagement. This usage serves to prevent digressions, de-escalate conflict, avoid sensitive issues, or save time, thereby maintaining conversational flow and coherence. As illustrated in (16), *hālā* enables speakers to gracefully exit a topic or signal disengagement with the current turn. By opting to sideline a new subject in favor of prior discourse, the speaker preserves coherence and avoids potential digressions. While this usage may appear to violate Grice's (1975) maxims (relevance, quantity, quality), it functions as a strategic communicative move to manage speaking time and mitigate conflict without disrupting the conversation.

(16)

*Man dar Tehrān ʔezdevāǰ kard-am… ʔun.vaqt*
I in Tehran marriage do.PST-1SG... that.time
*mā hālā... ʔalbate hāsel-e ʔun ʔezdevāǰ do.tā bače dār-im...*
we now... of.course result-of that marriage two.CL child have.PRS-1PL...
*ʔunhā ʔalān dige bozorg=an.*
they now anymore grown.up=COP.PRS.3PL.
'I got married in Tehran... then, we now... of course, the result of this marriage, we have two children... they are now grown up and have graduated from the university.'

#### *4.1.4.2 Structural functions of hālā in Persian*

This section explains that *hālā* can act as a boundary marker to separate different parts of a discussion and also helps show how different ideas or sentences are logically connected to each other.

***4.1.4.2.1 Hālā as a boundary marker.*** *Hālā* is often used to mark boundaries between discourse segments, such as clauses, turns, or topics.

(17)

*feʔlan ke xošbaxtāne dars=am tamum šod...*
now that fortunately studies=PC.1SG finished become.PST-3SG
*hālā bāyad kār.āmuzi tamum še madrak=am*
now must internship finished SBJV.become.PST-3SG degree=my
*ro be-gir-am.*
OM SBJV-take.PRS-1SG.
'Now my studies are finished... now, I have to finish my internship, get my degree.'

*Hālā* is used here to mark a transition, creating a clear boundary between clauses that cover different phases of the speaker's life, specifically moving from finishing their studies to beginning an internship. We mainly observe *hālā* performing this boundary-marking role in narratives, separating distinct sections of recalled events or memories. It also can be used to shift between narrative and non-narrative segments to provide background or marginal information while narrating.

*Hālā* may also act as a quotative marker, demarcating discourse parts and ensuring smooth transitions for the listener between direct speech, reported thought, and quoted material. It is vital to discourse structure, clearly signaling shifts among these modes. In this capacity, *hālā* typically occurs utterance-initially within a turn, following a reporting verb (especially cognitive verbs) and preceding the reported material. (See Ghaderi, 2021 for *xob* 'well' and Ghaderi & Amouzadeh, 2021 for *bale* 'yes' on this function). An example with *hālā* and *xob* in (18) will illustrate this (Ghaderi, 2022 justifies the emphatic function of *na* 'no' in such contexts.). In (18), after the verb *motavajǰeh šodan* ('notice'), *hālā* serves to mark the boundary between the speaker's realization, which is a thought process, and the actual content of that realization, specifically a comparison between different cultures.

(18)

| *Baʔd* | *ʔādam* | *kam.kam* | *motavajǰeh* | *mi-š-e* | *ke... xob* | *na* | *hālā* |
|---|---|---|---|---|---|---|---|
| Then | person | little.by.little | aware | IPFV-become.PRS-3SG | that...well | no | now |

| *yek* | *čiz-ā=yi* | *belʔaxare* | *tu* | *farhang=e* | *ʔinā* | *hast-eš* |
|---|---|---|---|---|---|---|
| INDF | thing-PL=INDF | finally | in | culture=EZ | they | be.PRS-3SG |

'Then, a person gradually realizes that... well, no, now, there are some things in their culture that might be better than my culture.'

***4.1.4.2.2 Hālā as a relational marker.*** *Hālā* is also used to signal logical relationships between ideas, such as contrast, consequence, or addition. Here, it functions as a relational marker, helping to connect different parts of the discourse.

(19)

| *Hālādo* | *tā* | *čiz=e* | *nesbatan* | *xeyli* | *qašang* | *na-goft-am...* |
|---|---|---|---|---|---|---|
| Nowtwo | CL | thing=EZ | relatively | very | beautiful | NEG-say.PST-1SG... |

| *hālā* | *yeki* | *az* | *čiz-hā=ye* | *ǰāleb=tari* | *ro* | *be-g-am.* |
|---|---|---|---|---|---|---|
| now | one | of | thing-PL=EZ | interesting=more | OM | SBJV-say.PRS-1SG. |

'Now, I didn't say two relatively nice things... now, I say one of the more interesting things'

Here, *hālā* shows contrast between the speaker's previous unpleasant statements and the new interesting point they are introducing. While *hālā* does not encode contrast itself, it frequently appears in this context to underscore differences between situations, expectations, or perspectives.

While signaling relations, *hālā* comes in some special collocation (constructions) to help convey specific meanings. At the end of this part, we will consider some of these constructions.

***Hālā... vali...* (concessive structure):** The Persian collocation *hālā… vali…* commonly expresses concession. It functions by excluding something mentioned after *vali* ("but"), either entirely or partially, from a general set or whole or a claim previously introduced before *hālā*. For instance, in (20), *hālā* signals an exclusion of Thailand and Spain from the countries due to some specific characteristics. This structure is vital for discourse coherence and cohesion, much like English concessive markers such as *although* or *even though*.

(20)

*hālā har kodum azin kešvar-hā ke mi-r-i...*
now every which of these country-PL that IPFV-go.PRS-2SG...
*yek xāter-āt=e xāssi piš mi-y-ād*
a memory-PL-EZ special happen IPFV-come.PRS-3SG
*vali... ʔin Espāniā vo Tāiland ham ǰozv=e list=e man hast-and*
but... this Spain and Thailand also part=EZ list=EZ I be.PRS-3PL
*ke saʔy mi-kon-am hamiše dobāre be=ešun bargard-am.*
That trying IPFV-do.PRS-1PL always again to=them SBJV.return.PRS-1SG.
'I have very good memories in Thailand... now, whichever of these countries you go to, you go with someone, and special memories happen, but... Spain and Thailand are also on my list that I always try to return to.'

***Hālā če... če...*** and ***hālā har*** **(alternative structure):** The Persian expressions *hālā če[X]če[Y]* ; literally 'now either X or Y' and *hālā har[X]*, literally, 'now any X' serve to indicate that alternatives or members of a set are interchangeable, highlighting a lack of meaningful difference, such that the result will be the same no matter which option is chosen. For example, in (21) *hālā če… če…* illustrates a lack of difference between generations, thereby emphasizing a shared lack of commitment in both groups. An example like *hālā har kešvar* (now any country) in (22) demonstrates this by implying that Iranians' preference for speaking the official language remains consistent regardless of the specific country. Both of these *hālā* constructions are valuable when a speaker wants to minimize the importance of a choice or convey equivalence.

(21)

*nasl=e mā hālā če nasl=e dāxel=e Irān če*
generation=EZ our now whether generation=EZ inside=EZ Iran or
*nasl=e xāreǰ=e Irān... nasl=e mā nasl=e*
generation=EZ outside=EZ Iran… generation=EZ our generation=EZ
*pāyband=i nist.*
committed=IND NEG.be.PRS.3SG
[I think in our generation, an issue has arisen, and that is that] 'our generation, now, whether inside Iran or outside Iran, is not a committed generation.'

(22)

*tarǰih mi-d-an bā zabun=e rasmiy=e hālā har*
prefer IPFV-give.PRS-3PL with language=EZ official=EZ now every
*kešvar=i ke dar=eš hast-an bā ham sohbat bo-kon-an.*
country=INDF that in=PC3.SG be.PRS-3PL with each.other talk SBJV-do.PRS-3PL.
'Sometimes, Iranians outside of here prefer to speak in the official language, now, whatever country they are in, with each other.'

Section 4.1 has presented various functions of *hālā*. In the next part, the article will examine the prosodic features correlated with these functions.

## *4.2 Prosodic analysis of hālā*

Before examining how discourse function relates to the acoustic properties of *hālā* tokens, we first report the descriptive statistics of all acoustic measures, as shown in Table 2, then we conduct a correlation analysis among the acoustic variables, as shown in Table 3. These steps help us to assess whether any functional effects observed for a given acoustic measure reflect relatively independent, or whether they may partly reflect overlap with other correlated dimensions.

Descriptive statistics for all acoustic measures are presented in Table 2. The dataset includes measures of duration, pitch, and intensity, as well as their corresponding ranges.

Table 2. Descriptive statistics for acoustic measures of *hālā* tokens (N = 267).

| | Mean | SD | IQR | Range | N |
|---|---|---|---|---|---|
| **Duration_ms** | 245.235 | 94.818 | 111.47 | 692.05 | 267 |
| **Max_Intensity** | 71.89 | 5.523 | 6.995 | 28.293 | 267 |
| **Intensity_Range** | 15.819 | 7.55 | 11.39 | 34.357 | 267 |
| **Max_Pitch** | 209.82 | 90.499 | 105.932 | 498.459 | 267 |
| **Pitch_Range** | 49.979 | 74.085 | 30.389 | 484.567 | 267 |

The pairwise correlations presented in Table 3 show that duration is only weakly related to maximum pitch (r = 0.13) and maximum intensity (r = 0.09). The correlation between maximum intensity and maximum pitch is also weak (r = 0.08). By contrast, maximum pitch and pitch range are strongly correlated (r = 0.77) and maximum intensity and intensity range are moderately correlated (r = 0.21) which indicate substantial redundancy between these measures. Although pitch range shows a similarly low correlation with duration (r = 0.10), the difference relative to maximum pitch (r = 0.13) is minimal.

Table 3. Pairwise Pearson correlations among acoustic measures.

| | Duration_ms | Max_Intensity | Intensity_Range | Max_Pitch | Pitch_Range |
|---|---|---|---|---|---|
| **Duration_ms** | | 0.09 | 0.30 | 0.13 | 0.10 |
| **Max_Intensity** | | | 0.21 | 0.08 | 0.11 |
| **Intensity_Range** | | | | 0.04 | 0.12 |
| **Max_Pitch** | | | | | 0.77 |
| **Pitch_Range** | | | | | |

These correlational relationships suggest that duration, maximum pitch, and maximum intensity are the least correlated acoustic dimensions. For that reason, the regression analyses focus on these three measures to make it easier to interpret functional effects more independently and reduces redundancy among predictors. This choice is also consistent with previous work (Tang & Shaw, 2021), which treats maximum pitch and maximum intensity as primary measures of prosodic prominence.

### *4.2.1 Duration*

The initial duration model included all four functional predictors, Temporal, Textual, Modal, and Interactive, along with pre- and post-*hālā* speech rate as covariates. After stepwise model selection under maximum likelihood estimation, only Temporal and Textual were retained in the final model. Modal and Interactive did not improve model fit and were therefore excluded.

The trimmed final model is shown in Table 4. In this model, Textual function had a significant negative effect on log-transformed duration ($\beta = -0.067$, $p = 0.001$), which indicates that *hālā* tends to be shorter in textual uses. Temporal function showed a small positive effect, but this was not statistically significant ($\beta = 0.027$, $p = 0.126$), suggesting only a weak tendency for temporal uses to be longer. Both pre- and post-*hālā* speech rate also predicted duration. A higher pre-speech rate was associated with longer duration, while a higher post-speech rate was associated with shorter duration.

Overall, these results suggest that duration is a stable correlate of functional variation in *hālā*. Most clearly, textual uses are consistently realized with shorter duration.

Table 4. Linear mixed-effects model predicting log-transformed duration of *hālā* (n = 264)

| Predictor | β | SE | t | p |
|---|---|---|---|---|
| Intercept | 2.354 | 0.016 | 147.97 | < 0.001 *** |
| Pre-speech rate | 0.020 | 0.008 | 2.40 | 0.017 * |
| Post-speech rate | −0.046 | 0.008 | −5.55 | < 0.001 *** |
| Temporal (Tmp_c) | 0.027 | 0.017 | 1.53 | 0.126 |
| Textual (Textual_c) | −0.067 | 0.021 | −3.24 | 0.001 ** |

### *4.2.2 Pitch*

We next turn to pitch. After stepwise model selection, Interactive was the only functional predictor retained in the final pitch model. The trimmed final model is presented in Table 5.

Table 5. Linear mixed-effects model predicting log-transformed maximum pitch (n = 260)

| Predictor | β | SE | t | p |
|---|---|---|---|---|
| Intercept | 2.273 | 0.019 | 120.97 | < 0.001 *** |
| Pre-speech rate | 0.008 | 0.007 | 1.21 | 0.227 |
| Post-speech rate | 0.005 | 0.007 | 0.75 | 0.455 |
| Interactive (IntrF_c) | 0.019 | 0.014 | 1.37 | 0.173 |

In this model, Interactive function showed a small positive effect on log-transformed maximum pitch, but the effect was not significant (β = 0.019, p = 0.173). Neither pre- nor post-*hālā* speech rate significantly predicted pitch.

Overall, these findings suggest that pitch does not provide robust evidence for functional differentiation in the realization of *hālā*. Interactive uses show a slight tendency toward higher pitch, but the effect is weak and not statistically reliable in the trimmed model.

### *4.2.3 Intensity*

Intensity showed a somewhat different pattern. After stepwise model selection, Modal and Interactive were retained as functional predictors in the final model. The trimmed final model is presented in Table 6.

Table 6. Linear mixed-effects model predicting maximum intensity (dB) (n = 263)

| Predictor | β | SE | t | p |
|---|---|---|---|---|
| Intercept | 72.29 | 0.73 | 98.70 | < 0.001 *** |
| Pre-speech rate | −0.283 | 0.280 | −1.01 | 0.314 |
| Post-speech rate | −0.294 | 0.277 | −1.06 | 0.289 |
| Modal (ModF_c) | −1.076 | 0.570 | −1.89 | 0.060 (marginal) |
| Interactive (IntrF_c) | 1.665 | 0.589 | 2.83 | 0.005 ** |

In the trimmed model, Interactive function had a significant positive effect on maximum intensity (β = 1.665, p = 0.005), indicating that *hālā* tends to be produced with greater intensity in interactive uses. Modal function showed a negative effect that was only marginally significant (β = −1.076, p = 0.060), which suggests a weaker tendency for *hālā* to be produced with lower intensity in modal contexts. Pre- and post-*hālā* speech rate were not significant predictors.

These results indicate that intensity provides relatively stable evidence for functional differentiation. Interactive uses of *hālā* are consistently associated with higher intensity, while modal uses show a weaker and less robust tendency toward lower intensity.

### 4.2.4 Summary

Taken together, these results, summarized in Table 7, show a differentiated pattern of prosodic encoding. Duration emerges as the most stable and independent cue, particularly for textual functions, while intensity provides a secondary but reliable cue for interactive functions. In contrast, pitch does not show robust or consistent functional effects.

Table 7. Summary of functional effects across acoustic dimensions

| Function | Duration | Pitch | Intensity |
|---|---|---|---|
| Textual | shorter ** | — | — |
| Temporal | longer (n.s.) | — | — |
| Interactive | — | weak (n.s.) | higher ** |
| Modal | — | — | lower (marginal) |

## 4.3 Mediation analysis

To evaluate whether the functional effects observed in the main acoustic models could be explained by another acoustic dimension, we conducted a series of mediation analyses. The mediation models are summarized in Tables 8–10.

### 4.3.1 Duration as dependent variable

We first looked at duration to determine whether the functional effects remained once pitch and intensity were added to the model. As shown in Table 8, the negative effect of Textual function stayed significant in all of the trimmed duration models, and its coefficient changed only minimally. Temporal function, by contrast, remained non-significant throughout. Neither pitch nor intensity significantly predicted duration when added to the model. Pre- and post-*hālā* speech rate continued to be significant in all four models.

These results indicate that the shorter duration associated with textual uses of *hālā* is direct and stable pattern and cannot be explained by variation in pitch or intensity.

Table 8. Mediation models predicting log-transformed duration of *hālā*

| Predictor | Base Model | | | + Pitch | | | + Intensity | | | + Pitch + Intensity | | |
|---|---|---|---|---|---|---|---|---|---|---|---|---|
| | β | SE | p | β | SE | p | β | SE | p | β | SE | p |
| Intercept | 2.354 | 0.016 | < 0.001 | 2.218 | 0.127 | < 0.001 | 2.217 | 0.121 | < 0.001 | 2.123 | 0.161 | < 0.001 |
| Temporal (Tmp_c) | 0.027 | 0.017 | 0.126 | 0.027 | 0.017 | 0.127 | 0.026 | 0.017 | 0.135 | 0.026 | 0.017 | 0.135 |
| Textual (Textual_c) | −0.067 | 0.021 | 0.001 | −0.068 | 0.021 | 0.001 | −0.065 | 0.021 | 0.002 | −0.066 | 0.021 | 0.002 |
| Pitch | — | — | — | 0.060 | 0.056 | 0.283 | — | — | — | 0.050 | 0.057 | 0.377 |
| Intensity | — | — | — | — | — | — | 0.002 | 0.002 | 0.255 | 0.002 | 0.002 | 0.337 |
| Pre-speech rate | 0.020 | 0.008 | 0.017 | 0.019 | 0.008 | 0.021 | 0.020 | 0.008 | 0.015 | 0.020 | 0.008 | 0.018 |
| Post-speech rate | −0.046 | 0.008 | < 0.001 | −0.046 | 0.008 | < 0.001 | −0.046 | 0.008 | < 0.001 | −0.045 | 0.008 | < 0.001 |
| Number of observations | 264 | | | 264 | | | 264 | | | 264 | | |

### *4.3.2 Pitch as dependent variable*

We then turned to pitch. As Table 9 shows, Interactive function was not a significant predictor in any of the trimmed pitch models. Adding duration did not change that pattern in any meaningful way, and duration itself was also non-significant. A different result appeared when intensity was added: intensity emerged as a strong positive predictor of pitch. The same pattern was found in the model that included both duration and intensity. Intensity remained significant, while neither Interactive function nor duration reached significance.

These findings show that pitch variation in *hālā* is not robustly conditioned by discourse function. Instead, pitch seems to be more closely tied to intensity than to functional category or duration.

Table 9. Mediation models predicting pitch (log Hz)

| Predictor | Base Model | | | + Duration | | | + Intensity | | | + Duration + Intensity | | |
|---|---|---|---|---|---|---|---|---|---|---|---|---|
| | β | SE | p | β | SE | p | β | SE | p | β | SE | p |
| Intercept | 2.273 | 0.019 | < 0.001 | 2.106 | 0.108 | < 0.001 | 1.743 | 0.100 | < 0.001 | 1.689 | 0.131 | < 0.001 |
| Interactive (IntrF_c) | 0.019 | 0.014 | 0.173 | 0.019 | 0.014 | 0.176 | 0.009 | 0.014 | 0.488 | 0.010 | 0.014 | 0.479 |
| Duration (log10) | — | — | — | 0.072 | 0.045 | 0.115 | — | — | — | 0.028 | 0.043 | 0.516 |
| Intensity | — | — | — | — | — | — | 0.007 | 0.001 | < 0.001 | 0.007 | 0.001 | < 0.001 |
| Pre-speech rate | 0.008 | 0.007 | 0.227 | 0.006 | 0.007 | 0.353 | 0.018 | 0.006 | 0.003 | 0.017 | 0.006 | 0.005 |
| Post-speech rate | 0.005 | 0.007 | 0.455 | 0.009 | 0.007 | 0.226 | 0.005 | 0.006 | 0.419 | 0.007 | 0.007 | 0.333 |
| Number of observations | **260** | | | **260** | | | **259** | | | **259** | | |

### *4.3.3 Intensity as dependent variable*

Finally, intensity was examined as the dependent variable. As shown in Table 10, Interactive function remained a significant positive predictor across all of the trimmed intensity models. Modal function showed a negative effect in every model, although its strength varied somewhat: it was marginal in the baseline model, significant when duration was added, marginal again when pitch was added, and significant in the model that included both duration and pitch. Both duration and pitch were significant positive predictors of intensity whenever they were included.

Overall, these results show that intensity reflects both functional and acoustic influences. The positive association with Interactive function remains stable, while duration and pitch also contribute independently.

Table 10. Mediation models predicting intensity (dB)

| **Predictor** | **Base Model** | | | **+ Duration** | | | **+ Pitch** | | | **+ Duration + Pitch** | | |
|---|---|---|---|---|---|---|---|---|---|---|---|---|
| | **β** | **SE** | **p** | **β** | **SE** | **p** | **β** | **SE** | **p** | **β** | **SE** | **p** |
| Intercept | 72.285 | 0.732 | < 0.001 | 58.998 | 4.397 | < 0.001 | 57.055 | 4.402 | < 0.001 | 45.989 | 5.767 | < 0.001 |
| Modal (ModF_c) | −1.076 | 0.570 | 0.060 | −1.203 | 0.561 | 0.033 | −0.996 | 0.560 | 0.076 | −1.083 | 0.543 | 0.047 |
| Interactive (IntrF_c) | 1.665 | 0.589 | 0.005 | 1.649 | 0.578 | 0.005 | 1.721 | 0.585 | 0.004 | 1.794 | 0.567 | 0.002 |
| Duration (log10) | — | — | — | 5.683 | 1.854 | 0.002 | — | — | — | 5.091 | 1.795 | 0.005 |
| Pitch | — | — | — | — | — | — | 6.718 | 1.905 | < 0.001 | 6.334 | 1.852 | < 0.001 |
| Pre-speech rate | −0.283 | 0.280 | 0.314 | −0.431 | 0.279 | 0.124 | −0.125 | 0.265 | 0.638 | −0.194 | 0.260 | 0.456 |
| Post-speech rate | −0.294 | 0.277 | 0.289 | −0.015 | 0.286 | 0.959 | −0.387 | 0.277 | 0.163 | −0.105 | 0.281 | 0.710 |
| Number of observations | **263** | | | **263** | | | **263** | | | **262** | | |

### *4.3.4 Summary*

The mediation analyses show that the three acoustic dimensions do not pattern alike. Duration remains an independent cue, especially for textual function, since its main functional effect persists even when pitch and intensity are taken into account. Pitch, on the other hand, does not show a robust independent functional effect and appears to pattern more closely with intensity. Intensity occupies an intermediate position: it retains clear functional sensitivity, especially for interactive uses, but is also shaped by variation in duration and pitch.

### *4.4 Summary of functional and prosodic results*

The functional analysis reveals that *hālā* is a highly multifunctional item, operating across temporal, textual, interactive, and modal domains. These functions frequently overlap, with the majority of tokens serving multiple pragmatic roles simultaneously, a hallmark of advanced grammaticalization within the Discourse Grammar framework (Heine et al., 2013, 2020). Textual functions were the most frequent, followed by interactive and modal functions.

The prosodic analysis shows that these functional differences are reflected most clearly in duration and intensity. Textual uses are consistently shorter, while temporal uses show only a weak tendency toward longer duration. Interactive uses are associated with higher intensity, and modal uses show a weaker tendency toward lower intensity. Pitch, by contrast, does not emerge as a robust independent cue. Mediation analyses further confirm that the duration effect for textual uses is direct and independent, while intensity remains sensitive to both function and other acoustic dimensions.

Together, these findings demonstrate that *hālā*'s pragmatic versatility is systematically encoded in its acoustic realization, with duration and intensity serving as the primary prosodic correlates of functional differentiation.

## 5 General discussion

The findings demonstrate that Persian *hālā* has undergone extensive grammaticalization from a temporal adverb to a highly multifunctional discourse marker. Crucially, its diverse functions are systematically differentiated by prosody. These patterns align with cross-linguistic tendencies for temporal adverbs to evolve into theticals, while also revealing language-specific acoustic encoding strategies.

## *5.1 Grammaticalization: From temporal adverb to discourse marker*

The analysis highlights the advanced grammaticalization of *hālā* in Persian, demonstrating its cooption from a temporal adverb in sentence grammar to a thetical (primarily a discourse marker) in thetical grammar. This grammaticalization pathway aligns with established frameworks, particularly the stages proposed by Heine and Kuteva (2007, pp. 33–46; see also Heine et al., 2020; Traugott, 1982). The process begins with context extension, whereby a linguistic item acquires new meanings through use in novel contexts, leading to context-induced reinterpretation. These new functions include turn management, attention drawing, stance marking, and topic organization. Such extensions are deeply anchored in the immediate communicative moment, reflecting Hopper and Traugott's (2003) concept of 'persistence,' where the item's original characteristics continue to be mirrored in its evolving functions. The overlapping cases containing temporal and non-temporal uses of *hālā* supports this persistence: although over half of the tokens (127) showed temporal adverbial uses, purely temporal cases were rare (n = 12), as most temporal tokens also carried discourse marker functions.

With continued use across diverse contexts, pragmatic inferences became conventionalized, leading to new semantic codes. This paved the way for desemanticization, or semantic bleaching (Traugott, 1982; Wichmann, 2011), where *hālā* gradually shed or generalized its specific temporal meaning, acquiring more procedural and interactional significance. This phenomenon is evident when *hālā* functions in neutral interpersonal or textual roles, such as marking boundaries in narrating past events, where its original temporal sense is largely obscured.

Along with semantic changes, decategorization occurred, signifying the loss of *hālā*'s original morphosyntactic properties as a temporal adverb. Furthermore, phonetic erosion, specifically the reduced duration observed for textual uses of *hālā*, reflects ongoing phonological reduction associated with its grammaticalization in Persian, though this effect was not robustly attested across all discourse-marker functions.

The observed shift in *hālā*'s meaning from conceptual (propositional) to procedural aligns with core principles of grammaticalization. Within thetical grammar, ongoing debates exist regarding the precise evolutionary trajectory: some scholars propose that interactives evolve into modals (Heine, 2023, p. 182), while others suggest the reverse (Traugott, 1982, 1995). While the precise direction of evolution remains debated, scholars working within thetical grammar generally concur that textual functions constitute a common endpoint or shared functional space for both interactive and modal markers. This convergence highlights the fundamental interconnectedness of these categories in discourse-level grammar.

## *5.2 Multifunctionality: A prototype of discourse versatility*

*Hālā* emerges as a prototypical thetical (or interactive) element. Similar to other theticals, instances of *hālā* used as a DM are invariable, discourse-level units. These markers are typically syntactically and prosodically distinct from the main utterance, do not alter their utterance's core propositional meaning, and being indexical, derive their interpretation from the immediate communicative context, the 'here-and-now' (Levinson, 2006). Their most noticeable feature, however, is their multifunctionality.

The analysis of 267 *hālā* tokens revealed significant multifunctionality, with most tokens serving multiple pragmatic roles concurrently. Regarding functional distribution (with overlaps allowed), 127 tokens had a temporal function, 211 textual, 100 interactive, and 92 modal. Approximately 70% of tokens exhibited multiple functions: 113 tokens served two functions, 66 had three, and 6 encompassed all four, with only 82 tokens restricted to a single function. Notably, these figures are approximate, as functional overlap and occasional coding disagreements may affect exact counts.

The qualitative discovery of *hālā*'s main and sub-functions, alongside their quantitative distribution (Table 1), reveals that for instances of *hālā*, in order of frequency, textual functions (topic shifting, signaling relationships, boundary marking, attention guidance, topic introduction, topic emphasis) are the most frequent, followed by interactive functions (turn management, listener engagement, feedback regulation), and then modal functions (epistemic stance, emotional expression, attitudinal marking).

Given this high degree of functional overlap, prosodic cues play a crucial role in disambiguation. Duration and intensity allow speakers to signal functional prominence even when multiple discourse roles are simultaneously active.

### *5.3 Acoustic correlates: the sound of meaning*

The pragmatic functions of *hālā* are reflected in its acoustic realization, but not all cues contribute in the same way. The clearest and most stable duration effect is linked to textual function: in the trimmed model, textual uses were significantly shorter than other uses. By contrast, temporal uses showed only a weak tendency toward longer duration, and modal and interactive functions did not show reliable duration effects. This shows that duration mainly helps distinguish textual uses of *hālā*, rather than serving as a general marker of all functional differences.

Some individual modal or interactive tokens may still sound slightly longer in context. These cases can be interpreted in terms of reinforcement (Heine, 2023, pp. 300-303), a mechanism where phonetic or pragmatic substance is enhanced through strategies like reduplication, lengthening, or added prosodic weight. This reinforcement serves to imbue the utterance with additional layers of meaning, such as emphasis, heightened attention, or increased engagement.

Intensity emerged as a particularly robust acoustic correlate. Interactive uses of *hālā* exhibited significantly higher intensity, signaling prominence and active engagement with the interlocutor ($\beta = 1.665$, $p = 0.005$). This heightened intensity appears to be a stable, robust feature marking interactive functions, not merely a consequence of longer duration or higher pitch. Conversely, modal uses were characterized by lower intensity, although this effect was only marginal after trimming ($\beta = -1.076$, $p = 0.060$), suggesting a more subdued realization when speakers convey epistemic or emotional stances, which include silent, internal thinking processes.

The speech rate surrounding *hālā* also proved influential. Both preceding and succeeding speech rates were significant predictors of *hālā*'s duration (Pre-speech rate: $\beta = 0.020$, $p = 0.017$; Post-speech rate: $\beta = -0.046$, $p < 0.001$). Specifically, a faster speech rate following *hālā* correlated with a shorter duration of the token itself. This interdependence highlights how the speaker's immediate tempo and the phonetic context shape the articulation of *hālā*.

Further mediational analyses revealed that observed effects on pitch were largely attributable to intensity. This suggests that while pitch variation exists, intensity may function as a more primary and independent acoustic marker, especially for delineating interactive functions, rather than pitch alone. In summary, the acoustic profile of *hālā* is dynamically modulated by its pragmatic role, with intensity playing a pivotal role in signaling interactive prominence.

These interpretations are supported by the statistical modeling results, which show that the duration effect for textual function and the intensity effect for interactive function remain robust after trimming. This confirms that duration and intensity, rather than pitch, constitute the primary acoustic correlates of functional differentiation in *hālā*.

The mediation analysis further refines this picture by showing that the observed duration pattern is largely direct and independent, whereas pitch variation is largely mediated by intensity. In contrast, intensity is associated with both duration and pitch, suggesting a more integrated acoustic system.

Taken together, the results suggest a hierarchy of acoustic cues: duration, especially for textual differentiation, and intensity, especially for interactional functions, emerge as the most informative parameters, while pitch remains a comparatively weak and unstable correlate.

## *5.4 Discourse markers: indexicality and efficiency*

Discourse markers (DMs) are fundamentally indexical, their meaning deeply contingent on the immediate context, including participants (speaker and hearer) and the discourse context (preceding or upcoming text) (Schiffrin, 1987, pp. 322–327), besides their evolving communicative situation. Forms like *hālā* exemplify the trajectory from temporal deictics, which anchor language to a specific point in time, to discourse deictics, which index elements within the discourse itself. This involves linking ideas, signaling topic shifts, or indexing speakers and listeners. This transition from temporal to discourse deixis facilitates the listener’s navigation and comprehension of the discourse structure (Levinson, 2006).

The frequent initial positioning of DMs like *hālā* in utterances correlates with high contextual probability. This predictability facilitates probabilistic shortening (Tang & Bennett, 2018), leading to phonological reduction compared to their original temporal forms. The prosodic reduction observed in *hālā* when functioning as a textual DM is a clear manifestation of this tendency, driven by usage patterns favoring economy and specialized function.

## *5.5 Cross-linguistic parallels and universal tendencies*

Heine et al. (2019) have examined generalizations regarding the unidirectional evolution of grammatical forms and constructions, drawing on data from over 1,000 languages. The grammaticalization pathway of *hālā* resonates with this extensive cross-linguistic research, which indicates that temporal adverbs commonly evolve into thetical elements. These elements serve crucial functions such as signaling various relations (adversative, causal, conditional), marking narrative segmentation, facilitating discourse shifts, and foregrounding actions or topics (Heine et al., 2019, pp. 425, 428, 474).

The functional overlap between Persian *hālā* and English *now* in marking topic boundaries, epistemic stance, turn management, contrast, and highlighting urgency or importance further underscores these universal tendencies in discourse marking strategies, in addition to their multifunctionality (Aijmer, 1988; 2002, Chapter 2; Hirschberg & Litman, 1987, 1993). Both words demonstrate how a basic temporal reference can be extended and bleached to serve complex interactive and textual functions.

## *5.6 Cultural nuances and future directions*

While *hālā*’s multifunctionality and grammaticalization align with universal patterns, its use in Persian politeness and indirectness will reveal its unique cultural aspects. Future research should explore written corpora and dialects, and conduct comparative studies to clarify language-specific nuances around such items. Our initial observations suggested a correlation between *hālā*’s high frequency, grammaticalization, semantic abstraction, and prosodic reduction in textual functions (aligning with the probabilistic reduction hypothesis), though this didn’t hold as strongly for other uses. These findings highlight the need for deeper exploration into the effects of syntactic position and other non-prosodic factors, in addition to pragmatic function, on the realization of *hālā*.

# 6 Conclusion

Overall, the findings demonstrate that *hālā* has undergone extensive grammaticalization, resulting in a highly multifunctional DM whose functions are systematically reflected in its prosodic realization. Duration and intensity emerge as the most robust acoustic cues, while pitch plays a more limited role. These results contribute to our understanding of the interaction between discourse function and prosody, and highlight the importance of multifunctionality in the analysis of DMs.

# Acknowledgments

# CRediT authorship contribution

**SG:** Conceptualization, Investigation, Data Annotation, Pragmatic analysis, Prosodic analysis, Methodology, Writing – original draft preparation, review and editing.
**MA:** Methodology, Software, Formal analysis, Prosodic analysis, Validation, Visualization, Writing– original draft preparation, review and editing.
**KT:** Conceptualization, Methodology, Formal analysis, Prosodic analysis, Writing – review and editing, Supervision.

# Statements and Declarations

**Ethics**: The study uses the publicly available LDC Persian Conversational Telephone Speech Corpus. No human participants were directly recruited, and ethics approval was not required. This is stated in the manuscript.
**Consent to participate:** Not applicable. (No human participants were directly recruited.)
**Consent for publication:** Not applicable. (No personal information was disclosed.)
**Conflict of interest:** The author(s) declared no potential conflicts of interest with respect to the research, authorship, and/or publication of this article.

**Funding:** No funding was received for this study.

**Data availability statement:** The derived data and scripts necessary to reproduce the results presented are available in Open Science Framework repositories and will be made public upon publication. The repositories are currently private; reviewer access can be provided upon request to the corresponding author. The original speech data come from the LDC Multi-Language Conversational Telephone Speech Corpus 2011, Central Asian Group, and cannot be shared publicly because they are licensed by the Linguistic Data Consortium.
**Declaration of generative AI use:** OpenAI’s ChatGPT was used for language and grammar editing and to help develop, revise, debug, and check selected Python and R scripts used for forced-alignment processing, utterance extraction, acoustic measurements, and statistical analysis. Google Gemini was used to produce the preliminary automatic transcriptions, as described in the Methods section. All AI-assisted text, code, and transcriptions were reviewed, tested, corrected, and verified by the authors. The AI tools did not replace the manual pragmatic annotation, correction of the forced alignments and acoustic boundaries, interpretation of the statistical results, or the authors' final evaluation of the findings. The authors take full responsibility for the final manuscript, code, analyses, results, and conclusions.

# Appendices

## *Appendix A. Development of a Persian Forced Aligner*

A Persian forced aligner was developed from scratch using the Montreal Forced Aligner (MFA, version 2.2.17**;** McAuliffe et al., 2017**).** Specifically, a custom Persian acoustic model and a pronunciation dictionary were created and subsequently adapted to the conversational *hālā* corpus.

### *A.1. Pronunciation Dictionary Construction*

#### A.1.1. Base Lexical Resource

The pronunciation dictionary was based on the Persian lexical resource distributed in the CharsiuG2P repository, specifically the file *dicts/fas.tsv* (Zhu et al., 2022). The initial lexicon was expanded and reformatted for compatibility with MFA. Entries were Unicode-normalized, cleaned of formatting artifacts, and standardized. Multi-word items were converted to underscore-separated forms. IPA transcriptions were segmented into space-separated phones while preserving long vowels and affricates as single units. Pronunciation variants were retained as separate entries. After cleaning and normalization, the final dictionary contained 8,035 entries and 31 unique phonemic symbols.

#### A.1.2. Handling Short Vowel Ambiguity

Persian orthography does not explicitly encode short vowels, which creates ambiguity for automatic grapheme-to-phoneme (G2P) conversion. Automatic G2P generation produced multiple pronunciation variants, and training a fully reliable G2P model proved difficult due to vowel underspecification. To reduce pronunciation noise during acoustic model training, we adopted a hybrid strategy. First, we retained only utterances whose words were fully covered by the curated dictionary. Second, out-of-vocabulary (OOV) words from the target conversational dataset were manually transcribed in IPA and added to the dictionary. Automatically generated pronunciations were manually verified before inclusion. This approach improved phonological consistency and reduced alignment instability caused by vowel ambiguity.

### *A.2. Acoustic Training Corpus Preparation*

#### A.2.1. Source Datasets

Because no pretrained Persian acoustic model was available for MFA, a custom acoustic training corpus was assembled from two publicly available Persian speech datasets. The first was the Persian subset of Mozilla Common Voice 15.0, a multi-speaker read-speech corpus distributed under an Apache 2.0 license (Ardila et al., 2020; Reza2kn, 2023). The second was ManaTTS Persian**,** a single-speaker, studio-quality corpus released under a CC-0 license and originally developed for text-to-speech research (Qharabagh et al., 2025). The two datasets complemented each other in important ways: Common Voice provided speaker variability, whereas ManaTTS provided a large amount of clean and acoustically consistent speech. Taken together, they formed a strong basis for training a robust Persian acoustic model for forced alignment.

#### A.2.2. Transcript Normalization and Pairing

Transcripts were cleaned by removing Persian and English punctuation and normalizing whitespace. For MFA compatibility, each audio file was paired with a corresponding .txt file containing the cleaned transcription. Audio files without matching transcripts were removed to prevent alignment failures.
Files were then organized into per-speaker folders, preserving speaker identity for speaker-aware acoustic modeling.

#### A.2.3. Dictionary Coverage Filtering

To minimize pronunciation uncertainty, only utterances whose words were fully covered by the curated pronunciation dictionary were retained. Utterances containing OOV words were excluded from the acoustic training corpus. This strict filtering reduced pronunciation ambiguity and improved model stability.

#### A.2.4. Audio Standardization

All audio files were converted to 16 kHz mono WAV format, which is the standard input format for Kaldi (used internally by MFA). Original MP3 files were removed after successful conversion to ensure uniformity.

#### A.2.5. Automatic Audio Quality Filtering

To improve acoustic model reliability, automatic audio quality filtering was applied prior to training. Each file was analyzed using frame-level acoustic measures, including clipping rate (signal saturation), DC offset, spectral flatness (noise-like characteristics), zero-crossing rate, speech activity ratio, and signal-to-noise ratio (SNR). These measures were used to identify recordings with excessive noise, distortion, long silence regions, or abnormal acoustic properties. Files that did not meet predefined quality criteria were excluded from the training set to prevent degraded model performance.

After this filtering procedure, the final acoustic training corpus consisted of 105,115 utterances totaling 94 hours and 22 minutes of speech.

#### A.2.6. Acoustic Model Training

The acoustic model was trained from scratch using MFA 2.2.17, with speaker information preserved throughout corpus organization to enable speaker-aware modeling. The workflow followed a standard sequence: first, the corpus was validated to ensure audio-transcript consistency, proper dictionary formatting, and absence of OOV errors; next, a new Persian acoustic model was trained on the 94-hour curated training corpus; this was followed by alignment of the training data to generate initial time boundaries.

## *Appendix B. Utterance segmentation*

All instances of the lexical item *hālā* were identified based on the hand-corrected word-level segmentations. For each token of *hālā*, a point annotation was placed at the temporal midpoint of the word on a dedicated tier for each channel. These midpoint annotations were then used as reference points for extracting the surrounding utterance context.

Following the practice of Seyfarth (2014), an utterance is defined as a stretch of speech produced by a single speaker and delimited by pauses around the target token. In line with this definition, utterance boundaries were determined on the basis of pauses in the acoustic signal. Pauses were detected automatically using an energy-based approach. Silent intervals were defined as segments of reduced intensity with a minimum duration of 300 milliseconds, following Biron et al. (2021), who treat pauses longer than 300 ms as relevant boundary cues in automatic prosodic segmentation**.** The silence threshold was set relative to the overall loudness of each recording: any interval at least 16 dB lower than the recording's average intensity was treated as a pause. If no pauses were detected with this initial threshold, the threshold was relaxed gradually in 1 dB steps to account for differences in recording quality and speaker loudness.

For each *hālā* token, potential pause boundaries were searched for within a ±10-second window centered on the token midpoint. Pauses preceding the token were treated as candidates for the left boundary, and pauses following the token were treated as candidates for the right boundary. When more than one candidate was available, longer pauses (≥ 500 ms) were given priority, and the pause closest to the token was selected. If no suitable pause was identified during the initial pass, pause detection was repeated locally within the same window.

The utterance was then defined as the segment between the end of the selected left pause and the beginning of the selected right pause. To keep the segments comparable and ensure consistency across them, only utterances with a total duration of no more than 10 seconds were retained. Finally, all automatically extracted utterances were manually inspected in Praat to verify boundary accuracy and to ensure that each segment represented a coherent stretch of speech centered on the target token.

## *Appendix C. Utterance extraction and annotation workflow*

### *C.1. Handling of overlapping utterances*

In some cases, multiple *hālā* tokens occurred in close succession, which led to overlapping utterance intervals. To preserve all valid segments while maintaining a clear annotation structure in Praat, overlapping utterances were distributed across separate tiers so that no two intervals overlapped within the same tier. Additional tiers were created automatically when required.

### *C.2. Metadata annotation and token tracking*

Each extracted utterance was assigned a unique identifier and labeled according to channel (caller vs. callee). These identifiers were written into the TextGrid files as additional tiers, allowing each token to be traced consistently throughout subsequent processing steps.

### *C.3. Manual verification and orthographic annotation*

After automatic segmentation, all utterances were manually inspected in Praat. Each interval was checked in context to confirm the accuracy of the automatically determined boundaries, and placeholder labels were replaced with orthographic transcriptions. This step ensured that the extracted segments were coherent stretches of speech centered on *hālā* and that their textual representations were accurate.

### *C.4. Summary construction and audio export*

Following manual annotation, the updated TextGrids were processed to generate a structured summary file. For each *hālā* token, the corresponding utterance interval containing the token midpoint was identified, and relevant metadata, including temporal span, channel, tier name, and transcription, were extracted into a CSV file with one row per token-centered utterance. The corresponding audio segments were then automatically extracted from the original recordings and saved as individual WAV files with matching text files. These paired files formed the working dataset for later analysis.

## *Appendix D. Abbreviations and phonetic notations*

| **Glossing abbreviations according to the Leipzig glossing rules** | | **Phonetics notations** | |
|---|---|---|---|
| AUX | auxiliary | ā | low back unrounded vowel |
| CLF | classifier | a | low front unrounded vowel |
| EZ | *ezafe* ‘addition’(-e or –ye) | u | high back rounded vowel |
| IND | indefinite | o | mid back rounded vowel |
| IPFV | imperfective aspect | i | high front unrounded vowel |
| NEG | negative | e | mid central unrounded vowel |
| OM | object marker | ʔ | voiceless glottal stop |
| PRS | present tense | q | voiceless uvular stop |
| PST | past tense | x | voiceless velar fricative |
| PC | pronominal clitic | š | voiceless palatal fricative |
| PL | plural number | ž | voiced palatal fricative |
| PP | past participle | č | voiceless palatal affricative |
| SBJV | subjunctive mood | ǰ | voiced palatal affricative |
| S | singular | h | voiceless glottal fricative |
| 1 | first person | y | voiced palatal glide |
| 2 | second person | (…) | an omitted, unnecessary part due to space limitations. |
| 3 | third person | | |
| - | morpheme boundary | [ ] | peripheral guiding text that is translated but not glossed. |
| = | clitic boundary | | |
| | | - | an interruption at the end of a turn. |